\documentclass[sigconf, natbib=false]{acmart}
\AtBeginDocument{%
  \providecommand\BibTeX{{%
    \normalfont B\kern-0.5em{\scshape i\kern-0.25em b}\kern-0.8em\TeX}}}

\copyrightyear{2024}
\acmYear{2024}
\setcopyright{acmlicensed}\acmConference[HRI '24 Companion]{Companion of the 2024 ACM/IEEE International Conference on Human-Robot Interaction}{March 11--14, 2024}{Boulder, CO, USA}
\acmBooktitle{Companion of the 2024 ACM/IEEE International Conference on Human-Robot Interaction (HRI '24 Companion), March 11--14, 2024, Boulder, CO, USA}
\acmDOI{10.1145/3610978.3640670}
\acmISBN{979-8-4007-0323-2/24/03}

\usepackage{color,soul}
\usepackage{graphicx} 
\usepackage{wrapfig}
\usepackage{amsmath}
\usepackage{caption}
\usepackage[toc,title,page]{appendix}
\usepackage{float}
\usepackage{amsfonts}
\RequirePackage[
  datamodel=acmdatamodel,
  style=acmnumeric,
  ]{biblatex}\usepackage{mathtools}
\usepackage{placeins}
\usepackage{chngcntr}
\usepackage{hyperref}
\begin{document}

\title{Kinematically Constrained Human-like Bimanual Robot-to-Human Handovers}

\author{Yasemin Göksu}
\authornote{The authors contributed equally to this research.}
\affiliation{
   \institution{Technische Universität Darmstadt}
   \city{Darmstadt}
   \country{Germany}
 }
 \email{goksuyasemin2@gmail.com}
 
\author{Antonio De Almeida Correia}
\authornotemark[1]
\affiliation{
   \institution{Technische Universität Darmstadt}
   \city{Darmstadt}
   \country{Germany}
 }
 \email{antonio.dealmeidacorreia@stud.tu-darmstadt.de}
 
\author{Vignesh Prasad}
\affiliation{
   \institution{Technische Universität Darmstadt}
   \city{Darmstadt}
   \country{Germany}
 }
 \email{vignesh.prasad@tu-darmstadt.de}
 
 \author{Alap Kshirsagar}
 \affiliation{
   \institution{Technische Universität Darmstadt}
   \city{Darmstadt}
   \country{Germany}
 }
 \email{alap.kshirsagar92@gmail.com}

\author{Dorothea Koert}
\affiliation{
   \institution{Technische Universität Darmstadt}
   \city{Darmstadt}
   \country{Germany}
 }
\email{dorothea.koert@tu-darmstadt.de}

\author{Jan Peters}
\affiliation{
   \institution{Technische Universität Darmstadt}
   \city{Darmstadt}
   \country{Germany}
 }
 \affiliation{
   \institution{German Research Center for AI}
   \city{Darmstadt}
   \country{Germany}
 }
  \affiliation{
   \institution{Hessian Center for AI}
   \city{Darmstadt}
   \country{Germany}
 }
\email{jan.peters@tu-darmstadt.de}

\author{Georgia Chalvatzaki}
\affiliation{
   \institution{Technische Universität Darmstadt}
   \city{Darmstadt}
   \country{Germany}
 }
\affiliation{
   \institution{Hessian Center for AI}
   \city{Darmstadt}
   \country{Germany}
 }
 \email{georgia.chalvatzaki@tu-darmstadt.de}
\renewcommand{\shortauthors}{Yasemin Göksu et al.}

\begin{abstract}
Bimanual handovers are crucial for transferring large, deformable or delicate objects. This paper proposes a framework for generating kinematically constrained human-like bimanual robot motions to ensure seamless and natural robot-to-human object handovers. We use a Hidden Semi-Markov Model (HSMM) to reactively generate suitable response trajectories for a robot based on the observed human partner's motion. The trajectories are adapted with task space constraints to ensure accurate handovers. Results from a pilot study show that our approach is perceived as more human--like compared to a baseline Inverse Kinematics approach. 
\end{abstract}

\ccsdesc[500]{Computing methodologies~Learning from demonstrations}
\ccsdesc[500]{Computer systems organization~Robotics}

\keywords{Human-Robot Interaction, Bimanual Manipulation}


\begin{teaserfigure}
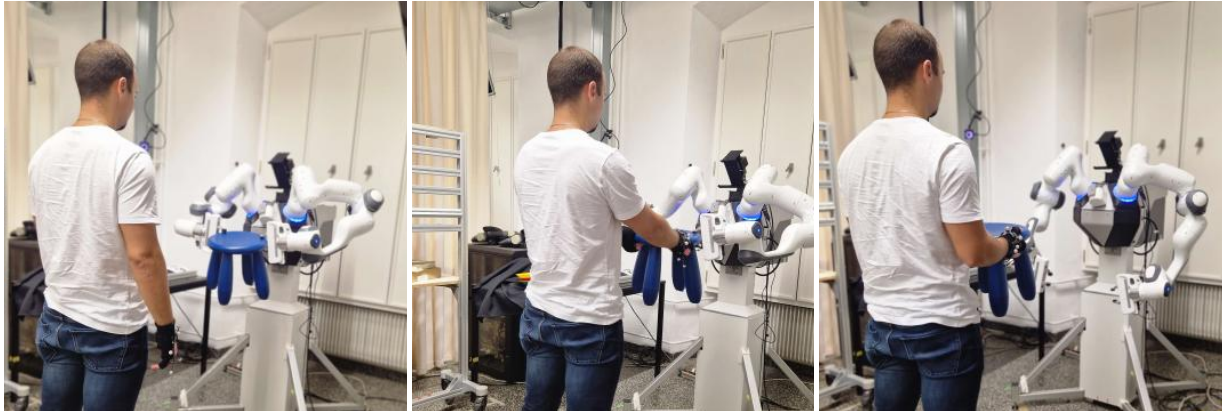

\begin{center}
      \includegraphics[trim={30 200 90 70},clip,width=.3\linewidth]{new_pics/handover0.pdf}
      \includegraphics[trim={30 200 90 70},clip,width=.3\linewidth]{new_pics/handover1.pdf}
      \includegraphics[trim={30 200 90 70},clip,width=.3\linewidth]{new_pics/handover2.pdf}
\end{center}
\caption{An example of the handover generated using our proposed pipeline. From left to right the pictures show the different stages of a handover motion where a robot hands over an object to a human partner. The robot motions are generated reactively by a Hidden-Semi Markov Model (HSMM) after observing the human partner's motion and are subsequently kinematically constrained to maintain the grip width, thereby ensuring a stable grasp during the handover.}
\label{fig:teaser}
\end{teaserfigure}


\maketitle

\section{Introduction}
In recent years, the significance of human-robot collaboration has grown, driven by the integration of robots into daily life~\cite{ajoudani2018progress}. A crucial aspect of such collaboration is the ability of robots to execute natural and intuitive object handovers with humans, embodying human-like behavior for seamless interactions~\cite{strabala2013toward}. While existing research primarily focuses on handovers of objects that require a unimanual grasp~\cite{ortenzi2021object, medina2016human, kshirsagar2022timing, iori2023dmp}, the need for bimanual handovers arises in scenarios involving large rigid items, deformable objects, spherical objects, and cultural etiquette. Surprisingly, there is limited research on robot controllers for bimanual handovers~\cite{ovur2023naturalistic, he2021bidirectional}, and the existing approaches lack the generation of human-like robot motions. 

In this work, we tackle the challenge of achieving human-like handover motions in bimanual robot-to-human object handovers through a learning-by-demonstration approach and the application of task space constraints on robot trajectories. Leveraging recorded demonstrations of human-to-human object handovers, we train a Hidden Semi-Markov Model (HSMM) on time-dependent states from trajectories of different lengths. The trained model predicts robot hand trajectories, conditioned on observed human trajectory. To ensure a constant grip-width required in a bimanual handover, we subsequently adapt the predicted robot trajectories through convex optimization with predefined constraints. We conducted a pilot study comparing our proposed method to a baseline controller for bimanual robot-to-human handovers. In the study, participants performed multiple handovers with a bimanual robot arm setup and assessed the robot motion based on various criteria, affirming the potential for our methodology to yield more human-like behavior.

\begin{figure}[ht]
    \centering
    \includegraphics[width=0.47\textwidth]{"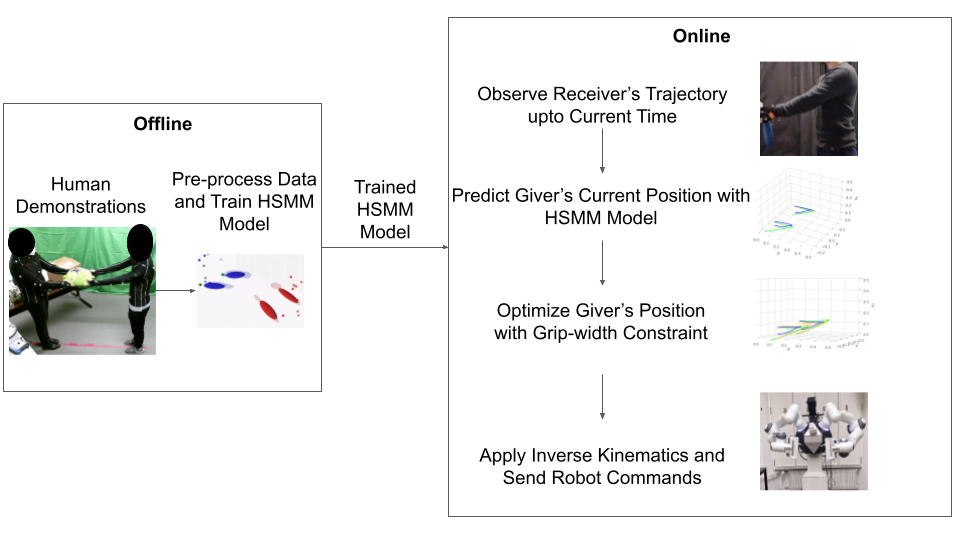"}
    \caption{This diagram gives an overview of our proposed pipeline. First, we train the HSMM model using recordings of human-to-human handovers. The trained HSMM model is used to predict robot hand trajectories during a robot-to-human handover. For each time step, we predict the giver's hand positions based on the receiver's hands' trajectory. We then optimize the predicted giver hand positions to maintain a constant grip-width. Finally, we generate the appropriate robot arm motion to reach the predicted position by applying inverse kinematics.}
    \label{fig:pipeline}
\end{figure}

From the predicted velocity of the giver's hands, we compute the target position of the giver's hands for the next timestep.

\section{Learning Kinematically Constrained Bimanual Handovers}
\label{method}
We aim to generate human-like bimanual handover trajectories for a robot-to-human object handover. Our proposed approach is shown in Fig.~\ref{fig:pipeline}. The first part of the process is done offline, which starts with the preprocessing of recorded human-human handover data to ensure that the giver’s and receiver’s trajectories are aligned properly. Subsequently, an HSMM is trained to predict the three-dimensional hand velocity of the giver based on the sequences of the following features: the receiver's hand positions, the receiver's hand velocities, and three relative distances between the giver's hand, the receiver's hand, and the object. During the execution on the robot, the trained model is used for predicting the robot giver's desired hand positions. Then convex optimization is applied to optimize the predicted positions obeying the grip-width constraint. The final step involves applying inverse kinematics to determine the robot's joint positions, which are sent to the robot as control inputs. 

In the rest of the paper, the term ``giver'' refers to the robot handing over the object, while the term ``receiver'' refers to the human receiving the object.


\subsection{Learning Bimanual Robot Handovers}
We train an HSMM to learn a joint distribution over the giver and receiver hand trajectories in a bimanual handover. We chose to employ HSMMs, since they are well suited for learning from demonstrations of different durations~\cite{s13253-021-00483-x, reader, segment}. An HSMM is a statistical model that can be applied over time series analysis problems and speech recognition. It is a variation of the hidden Markov model (HMM) \cite{bayes}. An HMM is defined by a set of hidden states $i\in\{1, 2, \dots, N\}$, where each one stands for a probability distribution. We use Gaussians with mean $\boldsymbol{\mu}_i$ and covariance $\boldsymbol{\Sigma}_i$, which characterize the emission probabilities of an observation $\mathcal{N}(\mathbf{y}_t;\boldsymbol{\mu}_i, \boldsymbol{\Sigma}_i)$. The initial probabilities of being in each state are represented by the initial state distribution $\pi_i$. We modeled three states that correspond to the reach, transfer, and retreat phases of a handover~\cite{kshirsagar2019specifying}. The probability of the model changing from the $i^{\text{th}}$ state to the $j^{\text{th}}$ state is described by the state transition probabilities $\mathcal{T}_{i,j}$. The forward variable of an HMM $h_i(\mathbf{y}_t)$ specifies the sequence of the components. 
\begin{equation}
\label{eq:hmm-h}
    h_i(\mathbf{y}_t) = \frac{\alpha_i(\mathbf{y}_t)}{\sum\nolimits_{j=1}^N\alpha_j(\mathbf{y}_t)}
\end{equation}
where
\begin{equation}
\label{eq:hmm-alpha}
\alpha_i(\mathbf{y}_t) = \mathcal{N}(\mathbf{y}_t;\boldsymbol{\mu}_i, \boldsymbol{\Sigma}_i)\sum_{j=1}^N\alpha_j(\mathbf{y}_{t-1})\mathcal{T}_{j,i}
\end{equation}
and $\alpha_i(\mathbf{x}_0) = \pi_i$. In the case of Hidden Semi--Markov Models, a Gaussian $p_i(d)$ is additionally fitted over the number of steps $d\in\{1,2,\dots D\}$ that the model stays in a given state, which then changes the forward variable as:
\begin{equation}
\label{eq:hsmm-alpha}
    \alpha_i(\mathbf{y}_t) = \mathcal{N}(\mathbf{y}_t;\boldsymbol{\mu}_i, \boldsymbol{\Sigma}_i)\sum_{j=1}^N\sum_{d=1}^Dp_i(d)\alpha_j(\mathbf{y}_{t-1})\mathcal{T}_{j,i}
\end{equation}

To encode the joint distribution between the interacting agents, we concatenate the features of both agents~\cite{calinon2009learning,evrard2009teaching} allowing the distribution to be decomposed as:
\begin{equation}
\label{eq:gmr}
    \boldsymbol{\mu}_i = \begin{bmatrix}
\boldsymbol{\mu}^1_i\\
\boldsymbol{\mu}^2_i
\end{bmatrix}; \boldsymbol{\Sigma}_i = \begin{bmatrix}
\boldsymbol{\Sigma}^{11}_i & \boldsymbol{\Sigma}^{12}_i\\
\boldsymbol{\Sigma}^{21}_i & \boldsymbol{\Sigma}^{22}_i
\end{bmatrix}
\end{equation}
where the superscript indicates the different agents.

The giver's features consist of the Cartesian velocity of the giver's hands, whereas the receiver's features consist of the Cartesian position and velocity of the receiver's hands. We construct three additional features: the distance of the giver's hands from the receiver's hands, the distance of the object from the giver's hands, and the distance of the object from the receiver's hands. We used these features in the model to distinguish the different phases of the handover interaction. The distances are considered relative to the distances at the middle of the transfer phase of each demonstration. Overall, the model is trained on trajectories from human-human demonstrations consisting of 15-dimensional values. Each phase gets a label to supervise the HSMM during training.

\subsection{Conditional Generation of Robot Trajectories}
We utilize the trained HSMM model to predict the velocity of the giver's hands for the current time step based on the input trajectory of the receiver's hands for prior time steps. To retrieve the giver velocities $\mathbf{y}^2_{1:t}$ given the receiver's observed trajectory $\mathbf{y}^1_{1:t}$, we use the formula
\begin{equation}
    \label{eq:gmr-conditioning}
    \mathbf{y}^2_{1:t} = \sum_{i=1}^N h_i(\mathbf{y}^1_{1:t}) (\boldsymbol{\mu}^2_i + \boldsymbol{\Sigma}^{21}_i(\boldsymbol{\Sigma}^{11}_i)^{-1}(\boldsymbol{\mu}^1_i - \mathbf{y}^1_{1:t})).
\end{equation}

We compute the predicted positions of the giver's hands by adding the predicted velocities to the hand positions from the previous timestep.

\subsection{Optimizing Predicted Trajectory}
The predicted positions of the giver's hands computed in the previous step do not ensure a constant grip width which may result in the object being stretched or crushed in a bimanual handover. To maintain the grip width while staying close to the predicted giver trajectory, we adapt the giver's hands' positions with convex optimization \cite{cvxbook} in which we minimize the squared euclidean distance between the optimized position and the predicted position. We model this using the following objective
\begin{equation}
   \mathrm{argmin}_\mathbf{x^{opt}_i} \lVert \mathbf{x}^{pred}_i-\mathbf{x}^{opt}_i\lVert^2,
\end{equation} where,
\begin{equation}
x^{opt}_i \coloneqq   (x^{opt}_{\mathrm{left}}, y^{opt}_{\mathrm{left}}, z^{opt}_{\mathrm{left}}, x^{opt}_{\mathrm{right}}, y^{opt}_{\mathrm{right}}, z^{opt}_{\mathrm{right}})_i^T,
\end{equation}
\begin{equation}
x^{pred}_i \coloneqq  (x^{pred}_{\mathrm{left}}, y^{pred}_{\mathrm{left}}, z^{pred}_{\mathrm{left}}, x^{pred}_{\mathrm{right}}, y^{pred}_{\mathrm{right}}, z^{pred}_{\mathrm{right}})_i^T
\end{equation}

are the optimized and predicted position vectors of the giver's hands at a given time $i \in {1,2,3,...}$. 


 
Given the initial hand positions of the giver ($x^{pred}_0 = x^{opt}_0$)  the initial grip width vector $g_0$ is defined as,


\begin{equation}g_0 =(g_{x}, g_{y}, g_{z})_0^T  \end{equation}\begin{equation*}\coloneqq(x^{pred}_{\mathrm{left}}, y^{pred}_{\mathrm{left}}, z^{pred}_{\mathrm{left}})_0^T - (x^{pred}_{\mathrm{right}}, y^{pred}_{\mathrm{right}}, z^{pred}_{\mathrm{right}})_{0}^T.
\end{equation*}

We define the grip width vector at any other timestep as,
\begin{equation}
g_i = T_ig_{i-1},
\end{equation}
where the rotation matrix $T_i$ fulfills the following equation 
\begin{equation}
(x^{pred}_{\mathrm{left}}, y^{pred}_{\mathrm{left}}, z^{pred}_{\mathrm{left}})_{i}^T - (x^{pred}_{\mathrm{right}}, y^{pred}_{\mathrm{right}}, z^{pred}_{\mathrm{right}})_{i}^T
\end{equation}\begin{equation*}= T_i((x^{opt}_{\mathrm{left}}, y^{opt}_{\mathrm{left}}, z^{opt}_{\mathrm{left}})_{i-1}^T - (x^{opt}_{\mathrm{right}}, y^{opt}_{\mathrm{right}}, z^{opt}_{\mathrm{right}})_{i-1}^T).
\end{equation*}


We use the following three grip width constraints corresponding to the three Cartesian axes:

\begin{equation}
   x^\mathrm{opt}_{\mathrm{left}}-x^\mathrm{opt}_{\mathrm{right}} = g_x, 
\end{equation}
\begin{equation}
   y^\mathrm{opt}_{\mathrm{left}}-y^\mathrm{opt}_{\mathrm{right}} = g_y, 
\end{equation}
\begin{equation}
   z^\mathrm{opt}_{\mathrm{left}}-z^\mathrm{opt}_{\mathrm{right}} = g_z, 
\end{equation}

The grip width constraints serve to ensure a secure object handover, preventing mishaps such as dropping (increased grip width) or compression (decreased grip width). \\Maintaining constant component--wise grip widths suffices for overall grip width consistency. We solve this convex optimization problem, involving the squared Euclidean distance and a linear function, using CVXPY \cite{CVXPY}. 

\subsection{Inverse Kinematics}
Given the optimized target position of the giver's hands obtained in the previous step, we compute the desired joint positions of the robot giver's arms using inverse kinematics~\cite{springer}. This involves solving a set of equations to determine the joint angles required to achieve the desired end--effector position. The joint angles are then sent to the low-level joint position controller of the robot. 


\section{Evaluation}
\label{experimentsAndResults}
We evaluate our proposed method in a study with human participants. The research question which we try to answer with our study is: ``To what extent is our kinematically constrained HSMM controller able to generate motion that is perceived as humanlike, compared to a baseline controller?''. The baseline controller produces robot hand trajectories that are directed towards the receiver's hand position at each timestep.

\subsection{Training Dataset and Experiment Setup}
We use a public dataset of human-to-human bimanual handovers~\cite{Dataset} to train the HSMM. Our training dataset includes the giver's and receiver's hand trajectories from 20 handovers of 10 different objects between two persons.


Our experiment setup consists of a bimanual robot and a set of cameras to track the hands of the participants. For the bimanual robot setup, we use two Franka-Emika Panda arms \cite{Franka} that are oriented such that they resemble the orientation of two human arms (see Fig.~\ref{fig:teaser}). We use an Optitrack motion capture system to track the hands of the human receiver.



\subsection{Study Design}
The independent variables for our study are the two controllers: our proposed kinematically constrained HSMM controller, and the baseline controller. The dependent variables are the average scores of parts of the Godspeed questionnaires~\cite{bartneckHRIMetrics2009}. We use the Anthropomorphism, Likeability, Perceived Intelligence, and Perceived Safety sections from the Godspeed questionnaire series. 
The participants receive three different objects from the robot: a folded cloth, a book, and children's stool.
Each person receives each object from the robot four times, twice with the baseline controller and twice with the kinematically constrained HSMM controller. The order of the controllers was decided according to the balanced Latin square. 
After performing handovers with each controller, the participant is asked to fill the questionnaire.  

\subsection{Experiment Results}
We conducted a pilot study with four participants to evaluate the human-likeness of our approach compared to the baseline controller. The participants (age 24--31) did not have prior experience of working with robots. Since each person tries every controller twice, we were able to collect eight scores per scale for the HSMM controller and eight scores per scale for the baseline controller. We used a one-sided paired t-test~\cite{doi:10.4097/kjae.2015.68.6.540} in order to assess whether the HSMM controller is able to produce higher Likert scores, compared to the baseline controller. 

\begin{table}
\begin{tabular}{|cccc|}
\hline
Scale                        & \begin{tabular}[c]{@{}c@{}}HSMM \\ Median\end{tabular} & \begin{tabular}[c]{@{}c@{}}Baseline \\ Median\end{tabular} & $p$-value                        \\ \hline
Unnatural to Natural         & 2.5                                                    & 2.0                                                        & 0.1753                           \\ \hline
Machinelike to Humanlike     & 2.0                                                    & 1.0                                                        & \textbf{0.0398} \\ \hline
Unconscious to Conscious     & 2.0                                                    & 2.0                                                        & 0.2584                           \\ \hline
Artificial to Lifelike       & 2.0                                                    & 2.0                                                        & 0.5                              \\ \hline
Moving Rigidly to Elegantly  & 3.0                                                    & 2.5                                                        & 0.0985                           \\ \hline
Dislike to Like              & 2.5                                                    & 2.5                                                        & 0.3423                           \\ \hline
Unfriendly to Friendly       & 3.0                                                    & 3.0                                                        & 0.3423                           \\ \hline
Unkind to Kind               & 3.0                                                    & 3.0                                                        & 0.6187                           \\ \hline
Unpleasant to Pleasant       & 3.0                                                    & 2.5                                                        & 0.0852                           \\ \hline
Awful to Nice                & 3.0                                                    & 2.5                                                        & 0.5                              \\ \hline
Incompetent to Competent     & 3.0                                                    & 2.5                                                        & 0.1753                           \\ \hline
Ignorant to Knowledgeable    & 3.0                                                    & 2.0                                                        & 0.0985                           \\ \hline
Irresponsible to Responsible & 2.5                                                    & 2.0                                                        & 0.0852                           \\ \hline
Unintelligent to Intelligent & 2.0                                                    & 2.0                                                        & 0.3423                           \\ \hline
Foolish to Sensible          & 3.0                                                    & 2.0                                                        & \textbf{0.0249} \\ \hline
Anxious to Relaxed           & 3.0                                                    & 2.5                                                        & 0.0518                           \\ \hline
Calm to Agitated             & 2.5                                                    & 2.0                                                        & 0.3341                           \\ \hline
Quiescent to Surprised       & 3.5                                                    & 3.0                                                        & 0.1425 \\ \hline                          
\end{tabular}
\caption{The median ratings of the two controllers and $p$-values of one-sided paired t-tests for the scales of the Godspeed Questionnaire Series}
\label{tab:results}
\end{table}

The results are shown in Table~\ref{tab:results}. We found that the HSMM controller gets perceived as more human--like rather than machine--like compared to the baseline controller (HSMM Median $2.0$, Baseline Median $1.0$, $p=0.0398$). Also, we found, that the HSMM controller gets perceived as more sensible rather than foolish compared to the baseline controller (HSMM Median $3.0$, Baseline Median $2.0$, $p=0.0249$). 

\section{Discussion}
\label{discussion}
We introduced a real--time kinematically constrained HSMM controller for robot-to-human bimanual object handovers. In a pilot study, we showed that our controller generated more human-like handovers compared to a baseline controller. However, the median rating of human-likenss was still low, which could be attributed to the machine-like appearance of the robot. Also, our proposed pipeline was not able to account for different heights of people, as people over the size of 1.85m generally criticized that they needed to bend over in order to retrieve the object. This could be the case, as the training data only involved two people, hence there was no height variance. Additionally, the effectiveness of our study was limited due to a very small sample size. But our pilot study showed promising results and we plan to conduct a full-scale study in the future.
\section*{Acknowledgements}
This work was supported by the German Research Foundation (DFG) Emmy Noether Programme (CH 2676/1-1), the German Federal Ministry of Education and Research (BMBF) Projects IKIDA (Grant no.: 01IS20045) and KompAKI (Grant no.: 02L19C150), the EU Projects MANiBOT and ARISE, and the Excellence Program, “The Adaptive Mind”, of the Hessian Ministry of Higher Education, Science, Research and Art.




\printbibliography

\end{document}